\documentclass[runningheads]{llncs}
\usepackage{graphicx}
\usepackage{booktabs}
\usepackage{tikz}
\usepackage{comment}
\usepackage{amsmath,amssymb} % define this before the line numbering.
\usepackage{color}
\usepackage{colortbl}
\usepackage[accsupp]{axessibility}  % Improves PDF readability for those with disabilities.

\usepackage[width=122mm,left=12mm,paperwidth=146mm,height=193mm,top=12mm,paperheight=217mm]{geometry}

\usepackage{multirow}
\usepackage{bm}
\usepackage{times,graphicx,amsmath,amssymb,booktabs,tabulary,multirow,overpic,xcolor}
\usepackage{amssymb}% http://ctan.org/pkg/amssymb
\usepackage{pifont}% http://ctan.org/pkg/pifont
\usepackage{booktabs}
\usepackage{color}
\definecolor{citecolor}{RGB}{34,139,34}
\definecolor{grayDark}{gray}{0.95}
\definecolor{grayLight}{gray}{0.98}
\definecolor{darkgreen}{rgb}{0.8, 0.1, 0.1}
\usepackage[pagebackref,breaklinks,colorlinks]{hyperref}

\usepackage[capitalize]{cleveref}
\crefname{section}{Sec.}{Secs.}
\Crefname{section}{Section}{Sections}
\Crefname{table}{Table}{Tables}
\crefname{table}{Tab.}{Tabs.}

\usepackage{bbm}

\begin{document}
% \linenumbers
\pagestyle{headings}
\mainmatter
\def\ECCVSubNumber{0000}  % Insert your submission number here

\title{Geometry-Guided Progressive NeRF for Generalizable \\ and Efficient Neural Human Rendering} % Replace with your title

% \begin{comment}
% \titlerunning{ECCV-22 submission ID \ECCVSubNumber} 
% \authorrunning{ECCV-22 submission ID \ECCVSubNumber} 
% \author{Anonymous ECCV submission}
% \institute{Paper ID \ECCVSubNumber}
% \end{comment}
%******************

% \begin{comment}
\titlerunning{GP-NeRF}

\author{Mingfei Chen\inst{1,2} \and
Jianfeng Zhang\inst{3} \and
Xiangyu Xu\inst{1} \and
Lijuan Liu\inst{1} \and
Yujun Cai\inst{1} \and
Jiashi Feng\inst{1} \and
Shuicheng Yan\inst{1}
}

\authorrunning{M. Chen et al.}
\institute{Sea AI Lab \and University of Washington \and National University of Singapore}
\maketitle

%%%%%%%%% ABSTRACT
\begin{abstract}
In this work we develop a \emph{generalizable} and \emph{efficient} Neural Radiance Field (NeRF) pipeline for high-fidelity free-viewpoint human body synthesis under settings with \emph{sparse} camera views. 
 % problem + previous
Though existing NeRF-based methods can synthesize rather realistic details for human body, they tend to produce poor results when the input has self-occlusion, especially for unseen humans under sparse views. Moreover, these methods often require a large number of sampling points for rendering, which leads to low efficiency and limits their real-world applicability.
 % our proposal
To address these challenges, we propose a Geometry-guided Progressive NeRF~(GP-NeRF).
In particular, to better tackle self-occlusion, we devise a geometry-guided multi-view feature integration approach that utilizes the estimated geometry prior to integrate the incomplete information from input views and construct a complete geometry volume for the target human body. 
% \textcolor{red}{(our target is improving generalization and efficiency, but here you mention you want to solve the self-occlusion problem.)}
Meanwhile, for achieving higher rendering efficiency, we introduce a  progressive rendering pipeline through geometry guidance, which leverages the geometric feature volume and the predicted density values to progressively reduce the number of sampling points and speed up the rendering process.
% experimental results
Experiments on the ZJU-MoCap and THUman datasets show that our method outperforms the state-of-the-arts significantly across multiple generalization settings, while the time cost is reduced $>70\%$ via applying our efficient progressive rendering pipeline.
% Code will be released. 
\end{abstract}
%%%%%%%%% BODY TEXT
%%%%%%%%% BODY TEXT
\section{Introduction}
\label{sec:intro}

% what task this paper aims at
High-fidelity free-viewpoint synthesis of human body is important for many applications such as virtual reality, telepresence and games.
Some recent works~\cite{park2020deformable,li2021neural,pumarola2021d,yuan2021star} deploy a Neural Radiance Fields~(NeRF)~\cite{mildenhall2020nerf} pipeline, which achieved fairly realistic synthesis of human body. 
However, these works usually require dense-view capturing of human body, and have to train a separate model for each person to render new views. The limited generalization ability as well as demand for cost computation severely hinder their application in the real-world scenarios.
\begin{figure}[t]
    \vspace{-2mm}
  \centering
\includegraphics[width=1.0\linewidth]{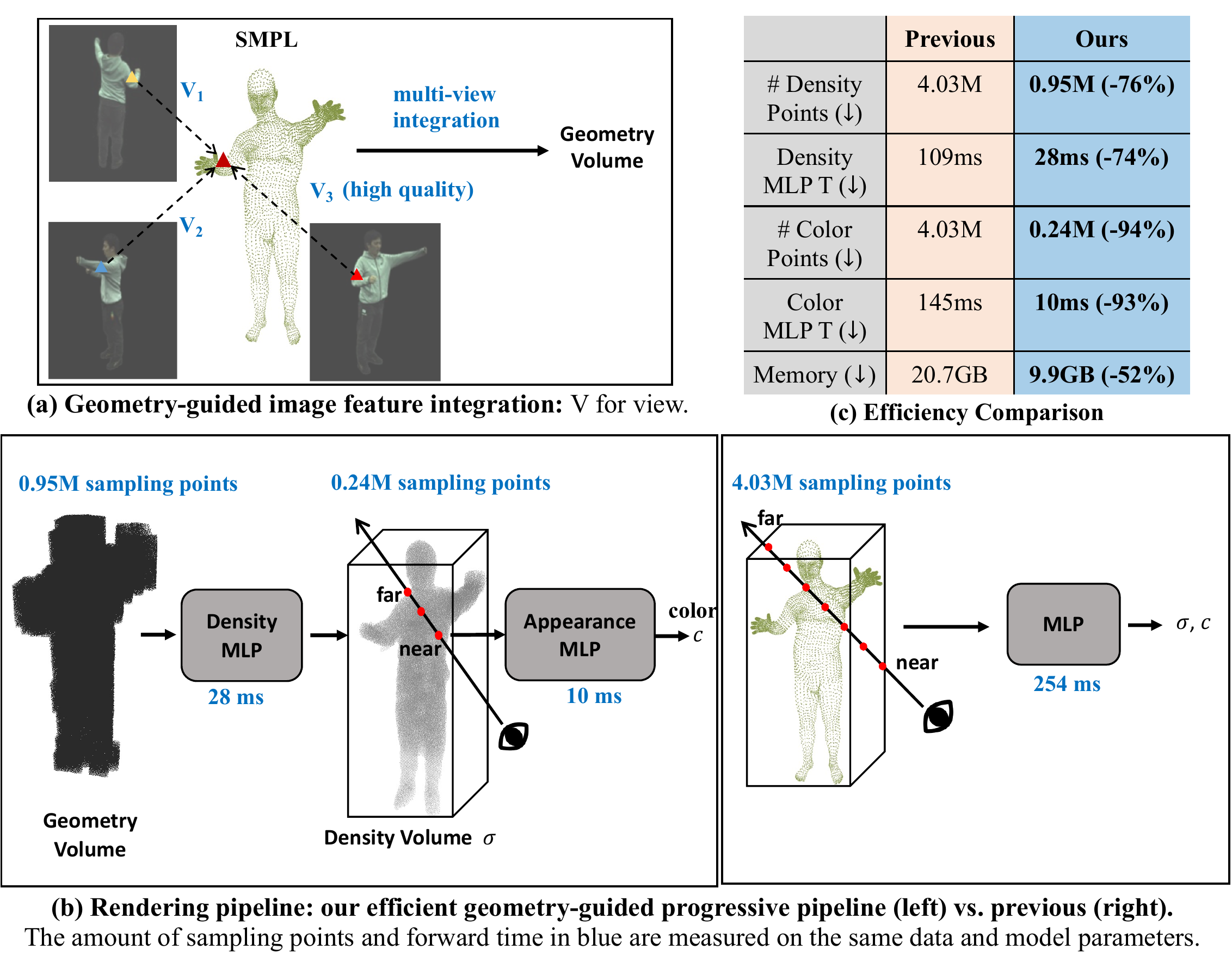}
\vspace{-5mm}
  \caption{Our method can better handle self-occlusion (a) and high computational cost (b) issues than previous methods~\cite{kwon2021neural,peng2021neural}.
  %that also utilize both geometry prior and image-conditioned features~\cite{kwon2021neural}.
  In (a), our multi-view integration can extract high-quality geometry information from $V_3$ for the red SMPL vertex. 
  In (b), our progressive rendering pipeline leverages the geometric volume and the predicted density values to progressively reduce the number of sampling points and speed up the rendering, while previous methods~\cite{kwon2021neural,peng2021neural} wastes large amount of computations at redundant empty regions. The efficiency comparison shown in (c) further verifies our high efficiency.
  } 
  \label{fig:intro}
  \vspace{-3mm}
\end{figure}
% % what is the research gap/challenge you are going to solve

In this work, we aim at boosting high-fidelity free-viewpoint human body synthesis with a generalizable and efficient NeRF framework based on only single-frame images from sparse camera views. 
To pursue such a high-standard framework, there are mainly two challenges that need to be tackled.
First, the human body is highly non-rigid and commonly has self-occlusions over body parts, which may lead to ambiguous results with only sparse-view captures.
This ambiguity could drastically degrade the rendering quality without proper regularizations, which cannot be easily solved by simply sampling features from multi-view images as in \cite{yu2020pixelnerf,wang2021ibrnet,raj2021pva}.
% or using multi-frame information from videos to compensate the occluded  information~\cite{peng2021neural,kwon2021neural}. 
This problem would become worse when using one model to synthesize unseen scenarios without specific per-scene training.
Second, the high computation and memory cost of NeRF-based methods severely hinder human synthesis with accurate details in high-resolution. 
For example, when rendering one $512\times 512$ image, existing methods need to process millions of sampling points through the neural network, even if using the bound of the geometry prior to remove empty regions.

% motivation of your proposed method to solve the research gap
To address these challenges, we propose a geometry-guided progressive NeRF, called GP-NeRF, for generalizable and efficient free-view human synthesis. 
%
%NeRF multi-view enhanced geometry prior, progressive rendering
More specifically, to regularize the learned 3D human representation, we propose a geometry-guided multi-view feature integration approach to more effectively exploit the information in the sparse input views.
For the geometry prior, we adopt a coarse 3D body model, \emph{i.e.}, SMPL~\cite{loper2015smpl}, which serves as a base estimate of our algorithm.
We attach multi-view image features to the base geometry model using an adaptive multi-view aggregation layer.
Then we can obtain an enhanced geometry volume by refining the base model with the attached image features, which substantially reduces the ambiguities in learning a high-fidelity 3D neural field.
It is worth noting that our multi-view enhanced geometry prior differs significantly from related methods that also utilize human body
priors~\cite{peng2021neural,kwon2021neural}.
NB~\cite{peng2021neural} learns a per-scene geometry embedding, which is hard to generalize to unseen human bodies; NHP~\cite{kwon2021neural} relies on temporal information to complement the base geometry model, which is less effective for regions occluded throughout the input video.
In contrast, our approach is able to adaptively combine the geometry prior and multi-view features to enhance the 3D estimation, and thus can better handle the self-occlusion problem and acquire lifted generalization capacity even without using videos (see Figure~\ref{fig:intro} (a)). By integrating the multi-view information and form a complete geometry volume adapting to the target human body, we can also compensate some limitations of the geometry prior~(e.g., inaccurate body shape or lacks cloth information), and support our following efficiency progressive pipeline well.

%In Figure~\ref{fig:intro} (a), we compare our integration approach with related methods that also use the geometry prior for human body synthesis~\cite{peng2021neural, kwon2021neural}. Instead of relying on temporal information to compensate the occluded information for each single view, we let each view to compensate the low-quality occluded information for other views with the guidance of the geometry prior. Our approach is more intuitive to handle the self-occlusion problem and gains even better generalization capacity simply based on the single frame.

Furthermore, to tackle the high computation and memory cost, we introduce a geometry-guided progressive rendering pipeline. As shown in Figure~\ref{fig:intro} (b), different from previous methods~\cite{peng2021neural,kwon2021neural}, our pipeline decouples the density and color prediction process, leveraging the geometry volume as well as the predicted density values to reduce the number of sampling points for rendering progressively.
By simply deploying our progressive rendering pipeline with the same data and model parameters, we can remove $76.4$\% points for density prediction~(with Density MLP in Figure \ref{fig:intro} (b)) and $94$\% points for color prediction~(with Appearance MLP in Figure \ref{fig:intro} (b)), reducing the total forwarding time of this part for all points by $85$\%. Later experiments verify that our progressive pipeline causes no performance decline while requiring shorter training time, which is credited to focusing on the density and appearance learning separately.

%  experiment performance and summary of the contribution
Our main contributions are in three folds: 
\begin{itemize}
    \item We propose a novel geometry-guided progressive NeRF~(GP-NeRF) for generalizable and efficient human body rendering, which reduces the computational cost of rendering significantly and also gains higher generalization capacity simply based on the single-frame sparse views.
    \item We propose an effective geometry-guided multi-view feature integration approach, where we let each view compensate the low-quality occluded information for other views with the guidance of the geometry prior.
    \item Our GP-NeRF has achieved state-of-the-art performance on the ZJU-MoCap dataset, taking only $175$ms on RTX $3090$ and reducing time for rendering per image by over $70\%$, which well verifies effectiveness and efficiency of our framework. 
\end{itemize}

%Besides the efficiency, we investigate how to improve the model's generalization capacity for synthesizing high-quality unseen scenarios like unseen poses or human bodies.
%
%First, with only sparse-view captures, the common self-occlusion over human body parts would drastically degrade the rendering quality, which cannot be easily solved by simply sampling features from multi-view images as in \cite{yu2020pixelnerf, wang2021ibrnet, raj2021pva} or using  multi-frame information from videos to compensate the occluded parts information~\cite{peng2021neural, kwon2021neural}.  
% main fig
%%%%%%%%% BODY TEXT
\section{Related Work}
\label{sec:related}

%quanhong: should explain why need review works on human performance capture
\paragraph{Human Performance Capture.} 
Previous works~\cite{newcombe2015dynamicfusion,collet2015high,dou2016fusion4d,guo2019relightables} apply traditional modeling and rendering pipelines for novel view synthesis of human performance, relying on either dense camera setup~\cite{debevec2000acquiring,guo2019relightables} or depth sensors~\cite{collet2015high,dou2016fusion4d,su2020robustfusion} to ensure photo-realistic reconstruction. 
%quanhong: check citation
Follow-up improvements are made by introducing neural networks to the rendering pipeline to alleviate geometric artifacts. 
To enable human performance capture in a sparse multi-view setup, template-based methods~\cite{carranza2003free,de2008performance,gall2009motion,stoll2010video} adopt pre-scanned human models to track human motion. However, these approaches require per-scene optimization and the pre-scanned human models are hard to collect in practice, which hinders them from real-world applications. 
Instead of performing per-scene optimization, recent methods~\cite{natsume2019siclope,saito2019pifu,saito2020pifuhd,zheng2019deephuman} adopt neural networks to learn human priors from ground-truth 3D data, and hence can reconstruct detailed 3D human geometry and texture from a single image. However, due to the limited diversity of training data, it is difficult for them to generate photo-realistic view synthesis or generalize to human poses and appearances that are very different from the training ones.

\paragraph{Neural 3D Representations.}
%quanhong: similarly, should explain why review this topic here, what is relation to your research
Recently, researchers adopt neural networks to represent the shape and appearance of scenes. These representations, such as voxels~\cite{sitzmann2019deepvoxels,lombardi2019neural,olszewski2019transformable,mescheder2019occupancy}, point clouds~\cite{aliev2020neural,wu2020multi}, textured meshes~\cite{liu2019neural,liao2020towards,liu2021spatt,xu2021texformer} and multi-plane images~\cite{flynn2019deepview,zhou2018stereo} are learned from 2D images via differentiable renderers. Though with impressive results, they are hard to scale to higher resolution due to innate cubic memory complexity. 

Researchers then propose implicit function-based approaches~\cite{sitzmann2019scene,liu2020neural,liu2020dist,niemeyer2020differentiable} to learn a fully-connected network to translate a 3D positional feature into local feature representation. A very recent work NeRF~\cite{mildenhall2020nerf} achieves high fidelity novel view synthesis by learning implicit fields of color and density along with a volume rendering technique. Later, several works extend NeRF to dynamic scenes modeling~\cite{park2020deformable,pumarola2021d,yuan2021star,li2021neural} by optimization NeRF and dynamic deformation fields jointly. 
Despite impressive performance, it is an extremely under-constrained problem to learn both NeRF and dynamic deformation fields together. NB~\cite{peng2021neural} combines NeRF with a parametric human body model SMPL~\cite{loper2015smpl} to regularize the training process.
It requires a lengthy optimization for each scene and hardly generalizes to unseen scenarios.
To avoid such expensive per-scene optimization, Generalizable NeRFs~\cite{raj2021pva,wang2021ibrnet,yu2020pixelnerf,kwon2021neural} condition the network on the pixel-aligned image features. However, directly extending such methods to complex and dynamic 3D human modeling is highly non-trivial due to self-occlusion, especially when modeling unseen humans under sparse views. Besides, these approaches suffer low efficiency since they need to process a large number of sampling points for volumetric rendering, harming their real-world applicability. Different from existing methods, we carefully design a multi-view information aggregation approach and a progressive rendering technique to improve model robustness and generalization to unseen scenarios under sparse views and also speed up the rendering.

% Unlike existing works, our method exploits temporal and multi-view information on-the-fly and achieves free-viewpoint human rendering in a \textit{feed-forward} manner, also generalizing to new, unseen human identities and poses. 

%%%%%%%%% BODY TEXT
\section{Methodology}
\label{sec:method}

% overview
Given a set of $M$ sparse source views \{$\mathbf{I}_m|m = 1, 2, ..., M$\} of an arbitrary human model, which are captured by $M$ pre-calibrated cameras respectively, we aim to synthesize the novel view $\mathbf{I}_t$ of the human model from an arbitrary target camera. 

\begin{figure*}[htb]
    \vspace{-3mm}
  \centering
\includegraphics[width=1.0\linewidth]{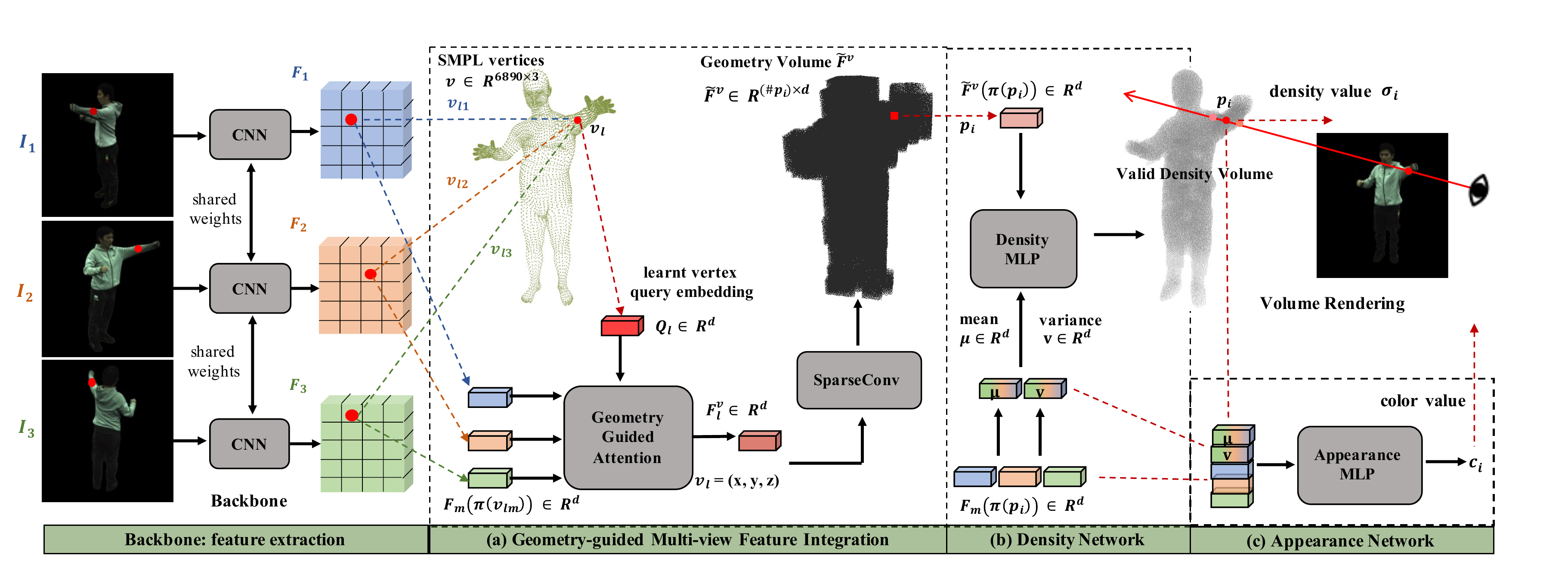}
  \caption{\textbf{Overview of our proposed framework.} Our progressive pipeline mainly contains three parts.  \textbf{(a) Geometry-guided multi-view feature integration.} We first learn query embedding $Q_l$ for each SMPL vertex to adaptively integrate the multi-view pixel-aligned image features $\mathbf{F}_m(\pi({v}_{lm}))$ through the geometry-guided attention module. Based on this, we utilize the SparseConvNet to construct a denser geometry feature volume $\tilde {\mathbf{F}}^v$. \textbf{(b) Density Network.} For point $p_i$ within $\tilde {\mathbf{F}}^v$, we concatenate its geometry feature $\tilde {\mathbf{F}}^v_i$ with the mean~($\bm{\mu}$) and variance~($\mathbf{v}$) of its pixel-aligned image features $\mathbf{F}_m(\pi(p_i))$, and predict its density value $\sigma_i$ through the density MLP. $p_i$ with a positive $\sigma_i$ form the valid density volume. \textbf{(c) Appearance Network.} For point $p_i$ within the valid density volume, we utilize $\mathbf{F}_m(\pi(p_i))$ to predict its color value $c_i$ through the appearance MLP.  Finally, we conduct the volume rendering to render the target image.}
  \label{framework}
  \vspace{-3mm}
\end{figure*}

To this end, we propose a geometry-guided progressive NeRF (GP-NeRF) framework for efficient and generalizable free-view human synthesis under very sparse views (e.g., $M = 3$). Figure \ref{framework} illustrates the overview of our framework.
Firstly,  a CNN backbone is used to extract image features $\mathbf{F}_m$ for each of the views $\mathbf{I}_m$. Then our GP-NeRF framework integrates these multi-view features to synthesize the novel-view image through three modules progressively, leveraging the geometry prior from SMPL~\cite{loper2015smpl}  as guidance. The three modules are 1) geometry-guided multi-view feature integration~(GMI) module~(Section \ref{GMI}); 2) density network (Section \ref{density}); and 3) appearance network~(Section \ref{color}). Details of the whole progressive human rendering pipeline are elaborated in Section \ref{HVR}, and the training method is described in Section \ref{training}.

% Geometry-guided Multi-view Integration
\subsection{Geometry-guided Multi-view Integration}
\label{GMI}
\vspace{-2mm}
The geometry-guided multi-view feature integration module, shown in Figure~\ref{framework} (a), 
enhances the coarse geometry prior with multi-view image features by adaptively aggregating these features via a geometry-guided attention module. Then it constructs a complete geometry feature volume that adapts to the target human body. %The process is detailed as follows.  

Firstly, we use the SMPL model~\cite{loper2015smpl} as the geometry prior, and get the pixel-aligned image features for each of the $6890$ SMPL vertices ${v}_l$ from each source image $\mathbf{I}_m$.
Specifically, we multiply the coordinate of ${v}_l$ with each source camera pose $[\mathbf{R}_m|\mathbf{t}_m]$ to transform the original ${v}_l$ to ${v}_{lm}$ into the source camera coordinate system, and then utilize the intrinsic matrix $\mathbf{K}_m$ to obtain the projected coordinate $\pi({v}_{lm})$ in the corresponding image plane.
We denote the pixel-aligned features from the image features $\mathbf{F}_m$ that corresponds to the pixel location of $\pi({v}_{lm})$ as $\mathbf{F}_m(\pi({v}_{lm}))$. We use bilinear interpolation to obtain the corresponding features if the projected location is fractional.

After obtaining $\mathbf{F}_m(\pi({v}_{lm}))$ from $M$ source views, we integrate them to represent the geometry information at vertex ${v}_l$ through a geometry-guided attention module.
Concretely, we learn an embedding $\mathbf{Q}_l$ for each ${v}_l$, and then take $\mathbf{Q}_l$ as a query embedding to calculate the correspondence score $s_{lm}$ with each $\mathbf{F}_m(\pi({v}_{lm}))$ respectively: 
\begin{equation}
\small
\begin{split}
s_{lm}^v= \frac {(\bm{W}_1 \mathbf{Q}_l+b_1)(\bm{W}_{2m} \mathbf{F}_{lm}^v+b_{2m})^{\top}}{\sqrt{d}}, \\ 
%& \mathbf{F}_{l}^v= \sum_{m=1}^{M} s_{lm}^v \mathbf{F}_{lm}^v
\end{split}
\end{equation}
where we denote $\mathbf{F}_m(\pi({v}_{lm}))$ as $\mathbf{F}_{lm}^v$ for simplicity. $d$ is the channel dimension of $\mathbf{F}_{lm}^v$. $\bm{W}$ represents linear projection layers.
After that, we weighted sum the $M$ pixel-aligned feature embeddings $\mathbf{F}_m(\pi({v}_{lm}))$ based on the scores $s_{lm}$ to obtain the aggregated geometry related feature $\mathbf{F}_{l}^v$ for vertex ${v}_l$:
%, $\mathbf{F}_{l}^v$ can be calculated as:
\begin{equation}
\small
\begin{split}
%& s_{lm}^v= \frac {(\bm{W}_1 \mathbf{Q}_l+b_1)(\bm{W}_{2m} \mathbf{F}_{lm}^v+b_{2m})^{\top}}{\sqrt{d}} \\ 
\mathbf{F}_{l}^v= \sum_{m=1}^{M} s_{lm}^v \mathbf{F}_{lm}^v.
\end{split}
\end{equation}

Considering the $6,890$ SMPL vertices with their corresponding features are not dense enough to represent the whole human body volume, we further learn to extend and fill the holes of the sparse geometry feature volume $\mathbf{F}^v = \{\mathbf{F}_{l}^v, l=1, 2, ..., 6890\}$ through the SparseConvNet~\cite{graham20183d} and thus obtain a denser geometry feature volume, denoted as $\tilde {\mathbf{F}}^v$. 
In our method, we take the geometry volume $\tilde {\mathbf{F}}^v$ as a more reliable basis to indicate occupancy of the human body in the whole space volume. More advanced than the coarse model SMPL, $\tilde {\mathbf{F}}^v$ leverages the multi-view image-conditioned features to enhance the coarse geometry prior, which adapts to the shape of the target human body. $\tilde {\mathbf{F}}^v$ only preserves the effective volume regions with body contents, including clothes regions. 
Because the SparseConvNet can gain experience from training to extend the features towards the regions with contents, based on the image-conditioned features with some instructive context information at each feature point.
Besides, the geometry volume will also benefit our progressive rendering pipeline, which will be detailed in Section~\ref{HVR}.

\vspace{-1mm}
% geometry related density network
\subsection{Density Network}\label{density}
\vspace{-2mm}
The density network predicts the opacity of each sampling point $\mathbf{p}_i$, which is highly related to the geometry of human body, like postures and shapes. Through the geometry-guided multi-view integration module in Section~\ref{GMI}, we can construct a geometry feature volume $\tilde {\mathbf{F}}^v$ which can provide sufficient reliable geometry information of the target human body. As shown in Figure~\ref{framework} (b), for each sampling point $\mathbf{p}_i$, we obtain its corresponding geometry related feature $\tilde {\mathbf{F}}^v_i$ from $\tilde {\mathbf{F}}^v$ based on its coordinate.
Though the feature volume can provide the geometry information of human body, such geometry-related features are coarse and may lose some fine image-conditioned features
that benefit the high-fidelity rendering. To compensate  the information loss, we combine these two kinds of features at the same sampling point to predict its density value more accurately.
Therefore, we further concatenate $\tilde {\mathbf{F}}^v_i$ with the mean~($\bm{\mu}$) and variance~($\mathbf{v}$) feature embedding of its corresponding pixel-aligned image features $\{\mathbf{F}_m(\pi({v}_{lm})), m=1, 2, ..., M\}$ that contain more detailed information, and process the concatenated feature through a density MLP to predict the density value at this point.

% appearance network
\vspace{-1mm}
\subsection{Appearance Network}\label{color}
\vspace{-2mm}
% TODO removing the view input or not?
The appearance network aims to predict the RGB color value for each sampling point $\mathbf{p}_i$. Since the RGB value is more related to the appearance details of human body, we utilize the image-conditioned features as the input to the appearance network for more detailed information.  As shown in Figure \ref{framework} (c), we first aggregate the pixel-aligned image features from input views for each color sampling point $\mathbf{p}^c_i$. Specifically, similar to obtaining the pixel-aligned image features for each SMPL vertex, we project the coordinate of $\mathbf{p}_i$ to the image plane of each source view, and obtain the pixel-aligned feature embedding, denoted as $\mathbf{F}_m(\pi(\mathbf{p}_{i}))$. We then concatenate $\mathbf{F}_m(\pi(\mathbf{p}_{i}))$ from $M$ source views with their mean~($\bm{\mu}$) and variance~($\mathbf{v}$) feature embeddings together. Afterwards, based on the concatenated feature embeddings, an appearance MLP is deployed to predict the RGB value $\hat{\mathbf{c}}_i = (\hat{r}_i, \hat{g}_i, \hat{b}_i)$ for the corresponding point $\mathbf{p}_i$.
% We assume that when two scenarios have different camera location settings or lighting conditions, the using of view directions for color prediction will have limited contributions to the generalization capacity across these two scenarios. Therefore, we remove the using of the view direction for color prediction, and focuses more on inferring the target color through the integrated information from the input views. Experiments in Section~\ref{mainresults} show that removing the view direction will not prevent ours from outperforming previous methods that use the view direction.

% geometry-guided progressive volume rendering 
\subsection{Geometry-guided Progressive Rendering}\label{HVR}
\vspace{-1mm}
% Afterwards, our geometry-guided progressive rendering pipeline utilizes the geometry volume to remove sampling points in the empty volumetric space for density prediction, and then refer to the predicted density values to further remove the redundant sampling points for the color prediction.
We render the human body in the target view through the volumetric rendering following previous NeRF-based methods~\cite{mildenhall2020nerf,peng2021neural,kwon2021neural}. Instead of sampling many redundant points for rendering, we introduce an efficient geometry-guided progressive rendering pipeline for the inference process. Our pipeline leverages the geometry volume in Section~\ref{GMI} as well as the predicted density values in Section~\ref{density} to reduce the number of points progressively. 

Specifically, we first preserve the sampling points that occupy the geometry volume $\tilde {\mathbf{F}}^v$ as valid density sampling points $\mathbf{p}_i^d$. 
Compared to the smallest pillar that contains the human body that is used by previous methods~\cite{peng2021neural,kwon2021neural}, the geometry volume is closer to the human body shape and contains much fewer redundant void sampling points. Then we predict the density values for $\mathbf{p}_i^d$ through the density network, and the sampling points that have positive density values form a valid density volume. As shown in Figure~\ref{framework}, the valid density volume is very close to the 3D mesh of the target human body and we further remove many empty regions compared to the geometry volume.
We take the sampling points in the valid density volume as the new valid sampling points $\mathbf{p}_i^c$, and further predict their color values through the appearance network in Section~\ref{color}.

We conduct volume rendering based on the density and color predictions to synthesize the target view $\mathbf{I}_t$. 
Traditional volume rendering methods often march rays $\mathbf{r}$ from the target camera to the pixels of the target view image, and then sample $N$ points on each $\mathbf{r}$. Denoting the distance of two adjacent sampling points on $\mathbf{r}$ as $\delta$, we can formulate the color rendering process for each $\mathbf{r}$ as:
\begin{equation}\label{eq:volume rendering}
\small
\begin{split}
& \hat{C}(\mathbf{r})=\sum_{i=1}^{N} T_{i}\left(1-\exp \left(-\sigma_{i} \delta_{i}\right)\right) \hat {\mathbf{c}}_{i}, \\ 
& \text{where}~T_{i}=\exp \left(-\sum_{j=1}^{i-1} \sigma_{j} \delta_{j}\right).
\end{split}
\end{equation}

For our progressive rendering pipeline, we use projection to bind the sampling points to $\mathbf{r}$. Concretely, we project the points within the geometry volume to the target view, take the nearest four pixels of the projected points as valid pixels to march a ray, and then uniformly sample $N$ points between its near and far bounds as~\cite{peng2021neural,kwon2021neural}. We only process the sampling points within the valid volume regions and then conduct volume rendering based on the rays $\mathbf{r}$.

Experiments in Section~\ref{efficiency} verify that our geometry-guided progressive rendering pipeline reduces the memory and time consumption during rendering significantly, and our performance can be even lifted by removing noisy unnecessary sampling points.

% training loss
\subsection{Training} \label{training}
During training, we do not deploy the progressive rendering pipeline in Section~\ref{HVR}, because it is useful only when our density network is reliable. Instead, we march rays from the target camera to pixels randomly sampled on the image while ensuring no fewer than half of the pixels are on the human body. 
We uniformly sample points on the rays to predict the corresponding density and color values. By performing the volume rendering in Eq.~(\ref{eq:volume rendering}), we obtain the predicted color $\hat{C}(\mathbf{r})$ for each $\mathbf{r}$. To supervise the network, we calculate the Mean Square Error loss between $\hat{C}(\mathbf{r})$ and the corresponding ground truth ${C}(\mathbf{r})$ color value as our training loss $\mathcal{L}_{rgb}$.

%%%%%%%%% BODY TEXT
\section{Experiments}
\label{sec:exp}
\vspace{-1mm}
We study four questions in experiments.
1) Is GP-NeRF able to improve the fitting and generalization performance of human synthesis on the seen and unseen scenarios~(Section \ref{mainresults})?
% 2) Can our method efficiently reduce the time and memory cost for rendering~(Section \ref{efficiency})?
2) Is GP-NeRF effective at reducing the time and memory cost for rendering~(Section \ref{efficiency})?
3) How does each individual design choice affect model performance~(Section \ref{ablations})
% 4) How can our visualization quality outperform other methods, both for human rendering and 3D reconstruction~(Section \ref{vis})?
4) Can GP-NeRF provide promising results, both for human rendering and 3D reconstruction~(Section \ref{vis})?
We describe the datasets and evaluation metrics in Section~\ref{metrics}, and our default implementation setting in Section~\ref{details}.

\definecolor{mygray}{gray}{0.6}
\definecolor{mygray-bg}{gray}{0.9}
\newcommand{\xmark}{\textcolor{red}{\ding{55}}}%
\newcommand{\cmark}{\textcolor{green}{\ding{51}}}%
\begin{table*}[h!]
\begin{center}
\caption{\textbf{Synthesis performance comparison}. Our proposed method outperforms existing methods on all the settings.}
\small
\resizebox{.8\textwidth}{!}{
\begin{tabular}{c|cc|c|cc|cc} %\hline
&\multicolumn{2}{c|}{Dataset} & Per-scene &\multicolumn{2}{c|}{Unseen} &  \multicolumn{2}{c}{Results} \\
Method   & Train & Test &  training  & Pose & Body & PSNR~($\uparrow$) & SSIM~($\uparrow$) \\ \hline
\multicolumn{8}{c}{Performance on training frames} \\
% \hline
  NT~\cite{thies2019deferred} &  ZJU-$7$&  ZJU-$7$ & \cmark      & \xmark & \xmark    & 23.86  & 0.896\\
NHR~\cite{wu2020multi} &  ZJU-$7$&  ZJU-$7$ & \cmark       & \xmark & \xmark    & 23.95  & 0.897      \\\
NB~\cite{peng2021neural} &  ZJU-$7$&   ZJU-$7$ & \cmark     & \xmark & \xmark      & 28.51 &  \textbf{0.947}  \\
NHP~\cite{kwon2021neural} &  ZJU-$7$&   ZJU-$7$ & \xmark      & \xmark & \xmark    & 28.73 &  0.936  \\ 
\cellcolor{mygray-bg}GP-NeRF~(Ours) &  \cellcolor{mygray-bg}ZJU-$7$&   \cellcolor{mygray-bg}ZJU-$7$ & \cellcolor{mygray-bg}\xmark      & \cellcolor{mygray-bg}\xmark & \cellcolor{mygray-bg}\xmark    &  \cellcolor{mygray-bg}\textbf{28.91} &  \cellcolor{mygray-bg}0.944 \\
% \cellcolor{mygray-bg}Ours &  \cellcolor{mygray-bg}ZJU-$7$&   \cellcolor{mygray-bg}ZJU-$7$ & \cellcolor{mygray-bg}\cmark     & \cellcolor{mygray-bg}\xmark & \cellcolor{mygray-bg}\xmark      & \cellcolor{mygray-bg} &  \cellcolor{mygray-bg}  \\
\hline
\multicolumn{8}{c}{Performance on unseen frames from training data} \\
NV~\cite{lombardi2019neural} &  ZJU-$7$&   ZJU-$7$ & \cmark   & \cmark & \xmark    & 22.00  & 0.818     \\
NT~\cite{thies2019deferred} &  ZJU-$7$&   ZJU-$7$ & \cmark   & \cmark & \xmark    & 22.28  & 0.872      \\
NHR~\cite{wu2020multi} &  ZJU-$7$&   ZJU-$7$ & \cmark   & \cmark & \xmark     & 22.31  & 0.871 \\
NB~\cite{peng2021neural} &  ZJU-$7$&   ZJU-$7$ & \cmark  & \cmark & \xmark     & 23.79 &  0.887  \\ 
NHP~\cite{kwon2021neural} &  ZJU-$7$&   ZJU-$7$ & \xmark  & \cmark & \xmark     & 26.94 &  0.929  \\ 
\cellcolor{mygray-bg}GP-NeRF~(Ours) &  \cellcolor{mygray-bg}ZJU-$7$&   \cellcolor{mygray-bg}ZJU-$7$ & \cellcolor{mygray-bg}\xmark      & \cellcolor{mygray-bg}\cmark & \cellcolor{mygray-bg}\xmark    &      \cellcolor{mygray-bg}{\textbf{27.92}} &  \cellcolor{mygray-bg}{\textbf{0.934}}\\
\hline
\multicolumn{8}{c}{Performance on test frames from test data} \\
NV~\cite{lombardi2019neural} &  ZJU-$3$&   ZJU-$3$ & \cmark   & \cmark & \xmark  & 20.84   &  0.827    \\
NT~\cite{thies2019deferred} &  ZJU-$3$&   ZJU-$3$ & \cmark   & \cmark & \xmark    & 21.92  & 0.873      \\
NHR~\cite{wu2020multi} &  ZJU-$3$&  ZJU-$3$ & \cmark   & \cmark & \xmark    & 22.03  & 0.875 \\
NB~\cite{peng2021neural} &  ZJU-$3$&   ZJU-$3$ & \cmark  & \cmark & \xmark    & 22.88 &  0.880  \\ 
PVA~\cite{raj2021pva} &  ZJU-$7$&   ZJU-$3$ & \xmark  & \cmark & \cmark    & 23.15 &  0.866  \\ 
Pixel-NeRF~\cite{yu2020pixelnerf} &  ZJU-$7$&   ZJU-$3$ & \xmark  & \cmark & \cmark     & 23.17 &  0.869  \\
% IBRNet~\cite{wang2021ibrnet} &  ZJU-$7$&   ZJU-$3$ & \xmark  & \cmark & \cmark     &  &    \\ 
NHP~\cite{kwon2021neural} &  ZJU-$7$&   ZJU-$3$ & \xmark  & \cmark & \cmark     & 24.75 &  0.906  \\
\cellcolor{mygray-bg}GP-NeRF~(Ours) &  \cellcolor{mygray-bg}ZJU-$7$&   \cellcolor{mygray-bg}ZJU-$3$ & \cellcolor{mygray-bg}\xmark      & \cellcolor{mygray-bg}\cmark & \cellcolor{mygray-bg}\cmark  &  \cellcolor{mygray-bg}{\textbf{25.96}} &  \cellcolor{mygray-bg}{\textbf{0.921}} \\
\hline
\multicolumn{8}{c}{Generalization performance across datasets} \\
NHP~\cite{kwon2021neural} &  AIST&   ZJU-$3$ & \xmark  & \cmark & \cmark      & 17.05 &  0.771  \\
% IBRNet~\cite{wang2021ibrnet} &  THuman-$7$&   ZJU-$3$ & \xmark  & \cmark & \cmark    &     &     &  &    \\ 
% \cellcolor{mygray-bg}Ours &  \cellcolor{mygray-bg}ZJU-$7$&   \cellcolor{mygray-bg}THuman-$7$ & \cellcolor{mygray-bg}\xmark      & \cellcolor{mygray-bg}\cmark & \cellcolor{mygray-bg}\cmark    &     \cellcolor{mygray-bg}&     \cellcolor{mygray-bg}&  \cellcolor{mygray-bg}&  \cellcolor{mygray-bg}  \\
\cellcolor{mygray-bg}GP-NeRF~(Ours) &  \cellcolor{mygray-bg}THUman-$7$&   \cellcolor{mygray-bg}ZJU-$3$ & \cellcolor{mygray-bg}\xmark      & \cellcolor{mygray-bg}\cmark & \cellcolor{mygray-bg}\cmark   &  \cellcolor{mygray-bg}24.74 &  \cellcolor{mygray-bg}0.907  \\
\cellcolor{mygray-bg}GP-NeRF~(Ours) &  \cellcolor{mygray-bg}THUman-all&   \cellcolor{mygray-bg}ZJU-$3$ & \cellcolor{mygray-bg}\xmark      & \cellcolor{mygray-bg}\cmark & \cellcolor{mygray-bg}\cmark    &    \cellcolor{mygray-bg}{\textbf{25.60}} &  \cellcolor{mygray-bg}{\textbf{0.917}}  \\
% \hline
\end{tabular}}
\end{center}
%\vspace{-2mm}

\label{tb:main}
%\vspace{-2mm}
\end{table*}
% dataset and metrics
\vspace{-1mm}
\subsection{Datasets and Metrics} \label{metrics}
\vspace{-1mm}
% zju + THuman
We train and evaluate our method on the ZJU-MoCap dataset~\cite{peng2021neural} and THUman $1.0$ dataset~\cite{zheng2019deephuman}. ZJU-MoCap contains $10$ sequences with $21$ synchronized cameras. We split the $10$ sequences into a training set with $7$ sequences and a test set with the remaining $3$ sequences, following~\cite{kwon2021neural} for a fair comparison. THUman contains $202$ human body $3$D scans. $80\%$ of the scans are taken as the training set, and the remaining are the test set. We render images for each scan from $24$ virtual cameras, which are uniformly set on the horizontal plane.

To evaluate the rendering performance, we choose two metrics: peak signal-to-noise ratio (PSNR) and structural similarity index (SSIM) following~\cite{mildenhall2020nerf,peng2021neural}. For the $3$D reconstruction, we only provide the qualitative results since the corresponding ground truth is not available.

\vspace{-1mm}
\subsection{Implementation Details} \label{details}
\vspace{-1mm}
In our implementation, we perform training and inference with an image size of $512 \times 512$ under $M=3$ camera views, where the horizontal angle interval is around $120^\circ$~(Uniform). We utilize a U-Net like architecture~\cite{wang2021ibrnet} as our backbone to extract the image features $\mathbf{F}$ in Section \ref{sec:method} with a dimension of $32$. 
%The ratio of marched rays from the target body mask regions is $0.5$ for training, and 
We sample $N=64$ points uniformly between the near and far bound on each ray. 
For training, we utilize the Adam optimizer~\cite{kingma2014adam}, and the learning rate decays exponentially from $1e-4$ for $180$k steps. We use one RTX $3090$ GPU with a batch size of $1$ for both training and inference.

\vspace{-1mm}
\subsection{Synthesis Performance Analysis} \label{mainresults}
\vspace{-1mm}
In Table~\ref{tb:main}, we compare our human rendering results to previous state-of-the-art methods. To evaluate the capacity of fitting and generalization on different levels, we train our framework on the first $300$ frames of $7$ training video sequences of ZJU-MoCap~(ZJU-$7$), and test on 1) the training frames, 2) unseen frames of ZJU-$7$, and 3) test frames from the $3$ test sequences~(ZJU-$3$), respectively. The results in Table~\ref{tb:main} verify our advanced generalization capacity on the unseen scenarios. We also achieve competitive fitting 
performance on the training frames, even comparable to the per-scene optimization methods~\cite{thies2019deferred,wu2020multi,peng2021neural}.

%\vspace{-2mm}
Notably, our method outperforms the state-of-the-art NHP~\cite{kwon2021neural} which utilizes the geometry prior with features of multi-view videos. Specifically, for the unseen poses and the unseen bodies, we outperform NHP by $0.98$ and $1.21$ dB on PSNR, and also by $0.5\%$ and $1.5\%$ on SSIM respectively, using only single-frame input. We also conduct generalization experiments across two datasets with different domains. We train our model on $7$ random human bodies from the THUman dataset~(THUman-$7$) and all $202$ human bodies~(THUman-all) separately, and test the synthesis performance on the test frames of ZJU-$3$. From Table~\ref{tb:main}, we observe our method outperforms NHP by a large margin under cross-dataset evaluation setup, i.e., around 7.7 dB and 13.6\% improvements on PSNR and SSIM respectively. All these results demonstrate the effectiveness of our geometry-guided multi-view information integration approach.

\vspace{-2mm}
\subsection{Efficiency Analysis} \label{efficiency}
\vspace{-2mm}
In Table~\ref{tb:speed}, we analyze the efficiency improvements
\footnote{We count averaged per-sample inference time in milliseconds. For all methods, the time is counted on NVIDIA GeForce RTX 3090 and CPU Intel i7-11700 @ 2.50GHz, PyTorch $1.8$, CUDA $11.4$.}
gained from our progressive pipeline on the first 300 frames of the 315~(Taichi) sequence in ZJU-MoCap dataset.

% time and memory cost
\begin{table}[h]
\centering

\caption{\textbf{Computation and memory cost comparison.} GP-NeRF$^{\dag}$ has the same structure as our GP-NeRF but adopts vanilla rendering technique. $\times N$ indicates the sampling points are split into $N$ chunks to be processed. \#$\mathbf{r}$ means the number of sampling rays; \#$\mathbf{p}^d$ and \#$\mathbf{p}^c$ mean sampling points through the density network and appearance network, respectively. T$^d$-total indicates the total time cost from backbone output to the density volume, including T$^d$-MLP which means the forwarding time of the density MLP. T$^c$-total means the time from density volume to the color prediction, and T$^c$-MLP is the time for the appearance MLP.}
\small
\resizebox{.95\textwidth}{!}{
\begin{tabular}{llllll}
% \hline
Method     &\#$\mathbf{r}$ (M)~($\downarrow$)     & \#$\mathbf{p}^d$ (M)~($\downarrow$) & \#$\mathbf{p}^c$ (M)~($\downarrow$) & Time (ms)~($\downarrow$) & Mem (GB)~($\downarrow$)  \\ \hline
NB 2$\times$~\cite{peng2021neural}  & 0.063 &   4.03     &   4.03     &   611     &     21.80     \\ % \hline
GP-NeRF$^{\dag}$ 3$\times$ & 0.063 (-0.0\%)  &  4.03  (-0.0\%)    &   4.03  (-0.0\%)   &   589 (-3.6\%)   & 14.53 (-33.3\%) \\
GP-NeRF$^{\dag}$ 2$\times$ & 0.063 (-0.0\%)  &  4.03  (-0.0\%)    &   4.03  (-0.0\%)   &   567 (-7.2\%)   & 20.74 (-4.9\%) \\
% Ours+ 3$\times$ & \bf 38.94 (-38.1\%)     &  \bf 0.95 (-76.4\%)    & \bf 0.24  (-94.0\%)   &   329.12 (-\%)  &    \bf 7.89  (-\%)    \\
GP-NeRF 2$\times$ & \bf 0.039 (-38.1\%)     &  \bf{0.95} (-76.4\%)   & \bf{0.24} (-94.0\%)   &   243 (-60.2\%)   & \bf{9.88  (-54.7\%)} \\
GP-NeRF 1$\times$ & \bf 0.039 (-38.1\%)     &  \bf{0.95} (-76.4\%)   & \bf{0.24} (-94.0\%)   & \bf{175} (-71.4\%) & {14.25} (-34.6\%) \\ % \hline
\\
%  \hline
% TODO this GPU?
Method           & T$^d$-MLP (ms)~($\downarrow$) & T$^d$-total (ms)~($\downarrow$) & T$^c$-MLP (ms)~($\downarrow$) & T$^c$-total (ms)~($\downarrow$) & PSNR~($\uparrow$) \\ \hline
GP-NeRF$^{\dag}$ 2$\times$      &   108.58   &  226.56    &  145.38   &   146.39  &    26.56     \\
% Ours 2$\times$      &   108.11~(-0.43\%)   &  215.37~(-4.9\%)     &  141.91~(-2.4\%)   &   142.92~(-2.4\%)  &    26.56~(+0.0\%)     \\
% Ours+ 3$\times$     &   32.26~(-70.3\%)   &  89.90~(-60.3\%)     &  10.70~(-92.6\%)   &   13.07~(-91.1\%)  & \bf 26.67~(+0.4\%)    \\
GP-NeRF 2$\times$     &   28.08~(-74.1\%)   &  83.65~(-63.1\%)     &  10.02~(-93.1\%)   &   11.4~(-92.2\%)  & \bf 26.67~(+0.4\%)    \\
GP-NeRF 1$\times$     &   \bf 23.55~(-78.3\%)   &  \bf 74.07~(-67.3\%)     &  \bf 9.50~(-93.5\%)    &   \bf 10.27~(-93.0\%)  & \bf 26.67~(+0.4\%)    % \\ \hline
\end{tabular}}
% \vspace{-2.5mm}

\label{tb:speed}
\end{table}

Considering the limited GPU memory, our final GP-NeRF can process all the sampling points in one run, but GP-NeRF$^{\dag}$ and NB~\cite{peng2021neural} requires at least twice.  As shown in the upper panel of Table~\ref{tb:speed}, compared to NB which also uses the SMPL bounds to remove redundant marched rays, our GP-NeRF can further remove $38.1\%$ rays and $76.4\%$ \#$\mathbf{p}^d$ by referring to the constructed geometry volume, and remove $94.0\%$ \#$\mathbf{p}^c$ based on the valid density volume. Correspondingly, our GP-NeRF achieves $175$ms per image for the whole rendering procedure, {$71.4\%$} less than NB and $70.3\%$ less than {GP-NeRF$^{\dag}$ $3\times$} which costs nearly the same GPU memory. For fair comparison to {GP-NeRF$^{\dag}$ $2\times$}, we also test the speed on {GP-NeRF} for $2$ chunks, and our progressive pipeline still reduces the time cost by $57\%$ and the memory cost by $52.4\%$, which verifies the significant efficiency improvement from the proposed rendering pipeline. 

In the bottom panel of Table~\ref{tb:speed}, we compare the time cost of each component in GP-NeRF to GP-NeRF$^{\dag}$ without progressive points reduction. The results show that we can reduce over $74\%$ and $63\%$ time cost for density MLP forwarding and the total density related time~T$^d$-total respectively, by simply using our progressive rendering pipeline on the same network structures. Our pipeline can also reduce over $92\%$ time cost for the appearance MLP forwarding. Moreover, our progressive pipeline improves the efficiency significantly while even improving the PSNR metric by $0.4\%$, as it can ignore some noisy sampling points during rendering that might degrade the performance.

% smpl-based aggregation
\begin{table}[ht]
\small
\begin{center}
\caption{\textbf{Ablations: feature integration.} G, Q, P are different approaches to obtain input features for the shared density and appearance network. G: geometry feature volume; Q: integrate multi-view information at each geometry vertex with geometry-guided attention; P: pixel-aligned image features.}
\begin{tabular}{l|ccc|cc}
% \hline
Variants & G & Q & P & PSNR~($\uparrow$) & SSIM~($\uparrow$)  \\
\hline
G & \cmark & \xmark & \xmark & 23.47 & 0.880\\
QG & \cmark & \cmark & \xmark & 23.68 & 0.885\\
P & \xmark & \xmark & \cmark & 26.09 & 0.915\\
QG+P & \cmark & \cmark & \cmark & \bf{26.69}& \bf{0.924}\\
% \hline
\end{tabular}
% \vspace{-2.5mm}

\vspace{-10mm}
\label{tb:feature ablations}
\end{center}
\end{table}

% disentangle entangle quick / long train
\begin{table}[hb]
\small
\begin{center}
\caption{\textbf{Ablations: progressive structure.} G, Q, P have the same meanings as Table~\ref{tb:feature ablations}. Disen. indicates whether the density~(Den.) and appearance~(App.) networks are in a progressive pipeline. Steps mean the number of training steps. The columns of Den. and App. demonstrate components of the input features.}
\begin{tabular}{cllc|cc}
% \hline
Disen. & Den. & App. & Steps & PSNR~($\uparrow$) & SSIM~($\uparrow$)  \\
\hline
\xmark & QG+P & QG+P & 5000 & 26.05 & 0.912\\
\cmark  & QG+P & QG+P & 5000 & 26.13 & 0.917\\
\cmark  & QG+P & P & 5000 & 26.16& 0.920\\ 
\cmark  & QG & P & 5000 & 25.71 & 0.904\\
\hline
\xmark & QG+P & QG+P & 35000 & 26.69 & 0.924\\
\cmark  & QG+P & QG+P & 35000 & 26.65 & 0.925\\
\cmark  & QG+P & P & 35000 & 26.67& 0.923\\
\cmark  & QG & P & 35000 & 26.40 & 0.918\\
\vspace{-10mm}
% \hline
\end{tabular}

\label{tb:disentangle ablations}
\end{center}
\end{table}

\vspace{-3mm}
\subsection{Ablation Studies} \label{ablations}
\vspace{-1mm}
We conduct ablation studies under the \textit{uniform camera setting} in Section \ref{details} to verify effectiveness of our main designed components on generalization capacity. We train our model on $7$ training sequences of the ZJU-MoCap dataset for $35$k steps and validate it on remaining 3 sequences.
% test the first $300$ frames of the remaining $3$ test sequences for validation.

\noindent\textbf{Feature Integration.}
In Table~\ref{tb:feature ablations}, we explore the effectiveness of the proposed geometry-guided feature integration mechanism on the baseline GP-NeRF, i.e., GP-NeRF without adopting progressive rendering pipeline.
% we discuss the geometry-guided feature integration mechanism on our baseline structure without a progressive pipeline, where the density and appearance network share the same network structure.
As shown in Table~\ref{tb:feature ablations}, adaptively aggregating multi-view image features with the guidance of the geometry prior to construct the geometry feature volume~(\textit{QG}) achieves better performance (i.e., $0.21$ dB and $0.5\%$ improvements on PSNR and SSIM respectively) than baseline that simply uses the mean of multi-view image features~(\textit{G}), as the proposed geometry-guided attention module helps focus more on the views corresponding to the geometry prior.
% Specifically, \textit{QG} gains improvements of $0.21$ dB and $0.5\%$  on PSNR and SSIM respectively. 
% Because through a learned geometry-guided attention module, we can focus more on the views that correspond more to the geometry prior, thus alleviating the self-occlusion problem.
We also observe baseline using only pixel-aligned image features (\textit{P}) gains $2.41$ dB PSNR and $3\%$ SSIM over baseline using only geometry feature (\textit{G}), as it captures more detailed appearance features from images for high-fidelity rendering.
% If we only use the image features \textit{P}, as shown in row $3$, we gain $2.41$ PSNR and $3\%$ SSIM over simply using the geometry feature volume that loses some detailed appearance-related features for high-fidelity rendering. 
Moreover, by combining the geometry feature and its corresponding detailed image features (\textit{QG+P}), we can improve upon \textit{P} by $0.6$ dB PSNR and $0.9\%$ SSIM respectively. This indicates that both the geometry and the pixel-aligned image features can compensate each other for better generalization performance on unseen scenarios.

% jianfeng: till here
\noindent\textbf{Progressive Structure.}
Our efficient progressive rendering pipeline in Section~\ref{HVR} requires a progressive structure of the density and appearance network. Based on the same experimental settings, we further decouple the density and appearance networks to form a progressive pipeline as in Figure~\ref{framework} and evaluate the performance. As shown in Table~\ref{tb:disentangle ablations}, the progressive structure does not harm the performance and even  reaches relatively high performance faster. This is because it allows these two networks to lean their different focus, thus improving the performance more quickly during training. 
For the density network, involving more detailed image features \textit{P} can enhance the relatively coarse geometry feature \textit{QG}, and bring around $0.5\%$ improvements on SSIM. The results also show that the geometry feature \textit{QG} is much more impactful on the geometry-related density prediction than on the appearance-related color value prediction. 
% \textcolor{red}{Considering training time, final generalization capacity and efficiency comprehensively, our final model deploys a progressive structure.}

% image visualization fig
\begin{figure}[!h]
    \vspace{-1mm}
  \centering
\includegraphics[width=0.9\linewidth]{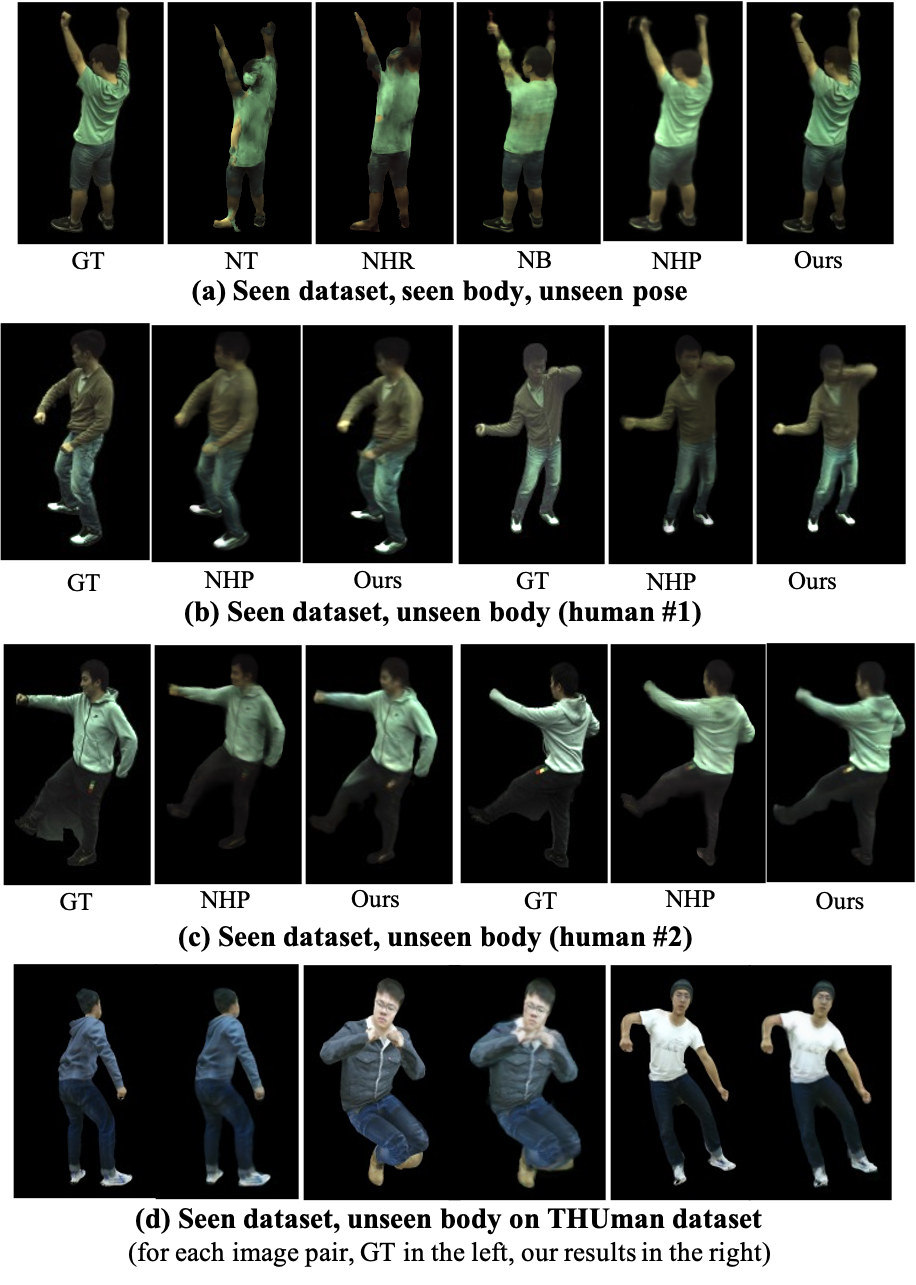}
  \caption{\textbf{Visualization comparisons on human rendering.} Comparing to other methods, ours can synthesize more high-fidelity details like the clothes wrinkles and reconstructing the body shape more accurately. Our synthesis can stick to the normal human body geometry better than methods without geometry priors like NT and NHR. We can also recover more accurate lighting conditions than the previous video-based generalizable method NHP on unseen bodies~(as (b) and (c)).}
  \label{image-visual}
  \vspace{-3mm}
\end{figure}

\vspace{-3mm}
\subsection{Visualization} \label{vis}
\vspace{-2mm}

We visualize our human rendering results under three uniform camera views in different experimental settings~(Figure~\ref{image-visual}). As Figure~\ref{image-visual} (a), (b) and (c) show, compared with other approaches, our method achieves better quality on unseen poses or bodies by synthesizing more high-fidelity details like the clothes wrinkles and reconstructing the body shape more accurately.  
From Figure~\ref{image-visual} (d), we demonstrate some rendering results on the unseen bodies of the THUman dataset after training on it. Our method generalizes well on the same THUman dataset and can synthesize accurate details. 
% Figure~\ref{image-visual} (d) demonstrates our generalization capacity across two datasets that have different domains. Trained on the THUman dataset, our method can still synthesize most details accurately on the unseen ZJU-MoCap dataset, even competitive to some previous methods trained on the ZJU-MoCap dataset.

%TODO analyze nhp: light condition, details...
%TODO analyze pifuhd, pixelnerf...

% mesh visualization fig
\begin{figure}[!ht]
    \vspace{-2mm}
  \centering
    \vspace{-3mm}
\includegraphics[width=0.8\linewidth]{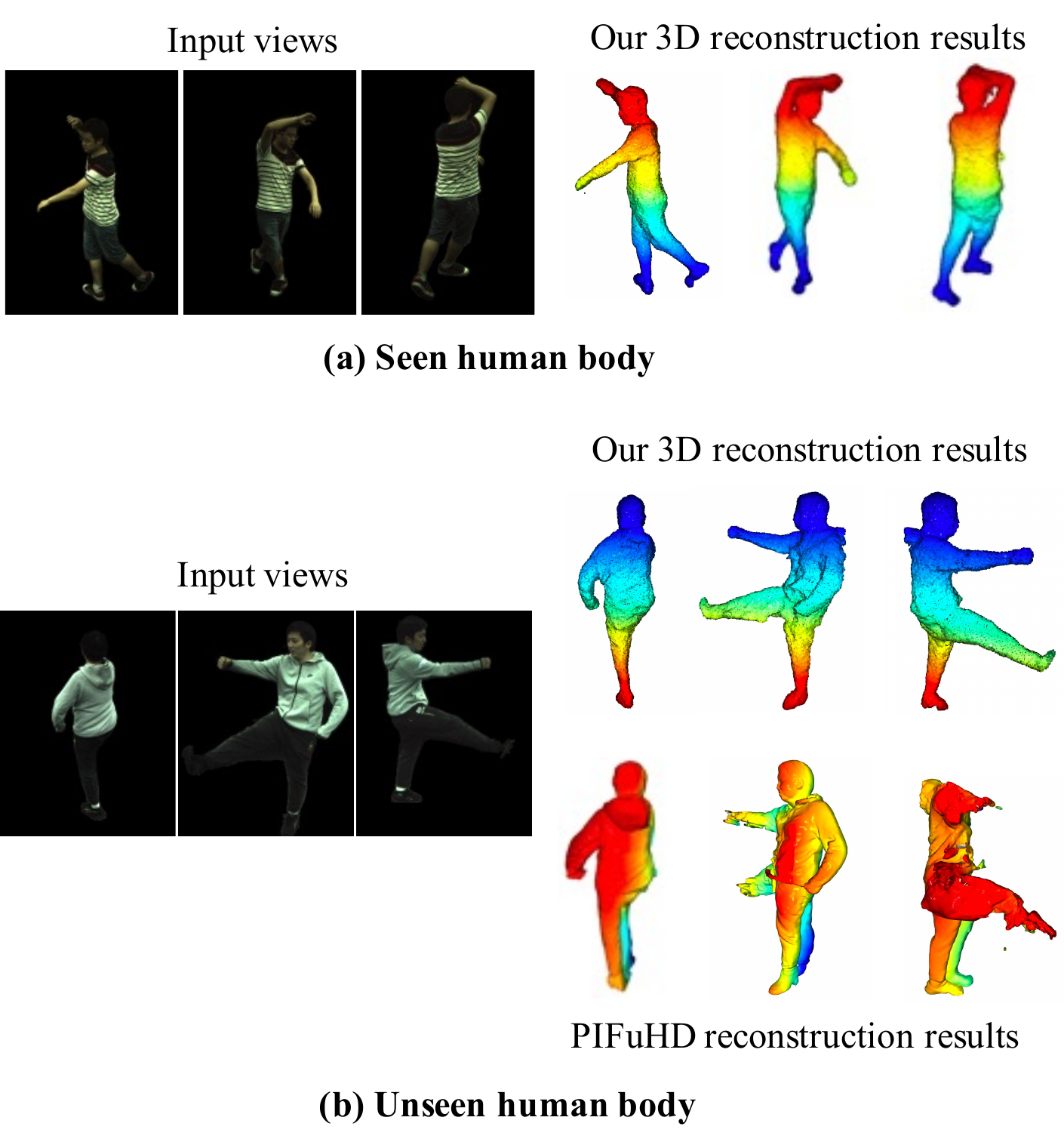}
  \caption{\textbf{Visualization of our 3D reconstruction results.} The color in the mesh is only for clearer visualization. By integrating multi-view information to form a complete geometry volume adapting to the target human body, our method can compensate some limitations of SMPL~(e.g., not accurate or lack cloth information), and can generally reconstruct very close human body shape and even clothes details like hoods and folds on unseen human bodies~(as (b)). We can generalize better on the unseen human bodies than previous image based 3D construction method like PIFuHD, which predicts incomplete or redundant body parts in its reconstruction results~(as (b)).}
  \label{mesh}
  \vspace{-4mm}
\end{figure}

In Figure~\ref{mesh} , we visualize the density volume from the density MLP in Section~\ref{density} as the mesh results of our 3D reconstruction.
Different from previous methods that densely sample points within bounds of the geometry prior to determine the inside points through the density network for mesh construction, our progressive pipeline directly determines the sampling points from the geometry volume in Section~\ref{GMI}, which contains much fewer redundant points and thus is more efficient for 3D reconstruction. Then we construct the mesh based on the points with higher density values. As Figure~\ref{mesh} (b) shows, on the unseen human bodies, previous image based 3D construction method like PIFuHD~\cite{saito2020pifuhd} can not generalize well. Besides their lower efficiency on making predictions for a lot of redundant sampling points, they are more likely to predict body parts that do not conform to a normal human body structure, because they can not integrate and adapt the given geometry information as well as we do. As shown in Figure~\ref{mesh}, by integrating multi-view information to form a complete geometry volume adapting to the target human body, our method can generally reconstruct very close human body shape and even clothes details like folds on even unseen human bodies~(Figure~\ref{mesh} (b)). 
% Our constructed mesh will lose some details on the unseen dataset because the generalization capacity degrades across different domains(e.g., with largely different human body shapes). However, we can still reconstruct plausible bodies, as shown in Figure~\ref{mesh} (c).

%%%%%%%%% BODY TEXT
\section{Conclusion}
\label{sec:conclusion}
We propose a geometry-guided progressive NeRF model for generalizable and efficient free-viewpoint human rendering under sparse camera settings.
Using our geometry-guided multi-view feature aggregation approach, the geometry prior can be effectively enhanced with the integrated multi-view information and form a complete geometry volume adapting to the target human body. The geometry feature volume combined with the detailed image-conditioned features can benefit the generalization performance on unseen scenarios.
We also introduce a progressive rendering pipeline for higher efficiency, which reduces over $70\%$ rendering time cost
% on the same data and module structures 
without performance degradation. 
Experimental results on two datasets verify our model can outperform previous methods significantly on generalization capacity and efficiency.

%%%%%%%%% REFERENCES

\clearpage
\bibliographystyle{splncs04}
\bibliography{egbib}
\end{document}